\documentclass{IEEEtran4PSCC}

\usepackage{booktabs}       
\usepackage{cite}

\ifCLASSINFOpdf
   \usepackage[pdftex]{graphicx}
\else
   \usepackage[dvips]{graphicx}
\fi
\usepackage[cmex10]{amsmath}
\usepackage{amsmath,amssymb,amsfonts}
\usepackage{enumitem}
\usepackage{float}
\usepackage{subcaption}
\usepackage{xcolor}

\usepackage[breaklinks]{hyperref}
\usepackage{url}
\usepackage{breakurl}

\floatstyle{ruled}
\newfloat{model}{thp}{lop}
\floatname{model}{Model}

\makeatletter
\let\old@ps@headings\ps@headings
\let\old@ps@IEEEtitlepagestyle\ps@IEEEtitlepagestyle
\def\psccfooter#1{%
    \def\ps@headings{%
        \old@ps@headings%
        \def\@oddfoot{\strut\hfill#1\hfill\strut}%
        \def\@evenfoot{\strut\hfill#1\hfill\strut}%
    }%
    \def\ps@IEEEtitlepagestyle{%
        \old@ps@IEEEtitlepagestyle%
        \def\@oddfoot{\strut\hfill#1\hfill\strut}%
        \def\@evenfoot{\strut\hfill#1\hfill\strut}%
    }%
    \ps@headings%
}
\makeatother

\psccfooter{%
        \parbox{\textwidth}{\hrulefill \\ \small{24th Power Systems Computation Conference} \hfill \begin{minipage}{0.2\textwidth}\centering \vspace*{4pt} \includegraphics[scale=0.06]{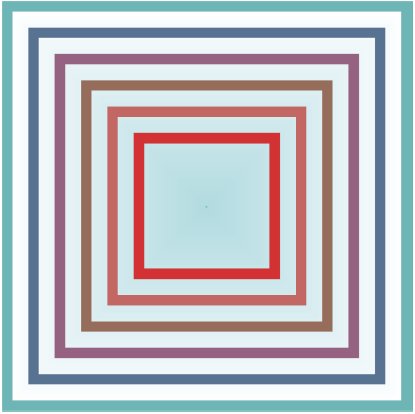}\\\small{PSCC 2026} \end{minipage} \hfill \small{Limassol, Cyprus --- June 8-12, 2026}}%
}

\begin{document}
%
\title{Machine Learning Guided Optimal Transmission Switching to Mitigate Wildfire Ignition Risk}

\author{
\IEEEauthorblockN{Weimin Huang\\ Bistra Dilkina}
\IEEEauthorblockA{Department of Computer Science \\
University of Southern California\\
Los Angeles, CA, USA\\
\{weiminhu, dilkina\}@usc.edu}
\and
\IEEEauthorblockN{Ryan Piansky\\ Daniel K. Molzahn}
\IEEEauthorblockA{School of Electrical and Computer Engineering\\
Georgia Institute of Technology\\
Atlanta, GA, USA\\
\{rpiansky3, molzahn\}@gatech.edu}
}


\maketitle

\begin{abstract}
To mitigate acute wildfire ignition risks, utilities de-energize power lines in high-risk areas. The Optimal Power Shutoff (OPS) problem optimizes line energization statuses to manage wildfire ignition risks through de-energizations while reducing load shedding. OPS problems are computationally challenging Mixed-Integer Linear Programs (MILPs) that must be solved rapidly and frequently in operational settings. For a particular power system, OPS instances share a common structure with varying parameters related to wildfire risks, loads, and renewable generation. This motivates the use of Machine Learning (ML) for solving OPS problems by exploiting shared patterns across instances. In this paper, we develop an ML-guided framework that quickly produces high-quality de-energization decisions by extending existing ML-guided MILP solution methods while integrating domain knowledge on the number of energized and de-energized lines. Results on a large-scale realistic California-based synthetic test system show that the proposed ML-guided method produces high-quality solutions faster than traditional optimization methods.

\end{abstract}


\thanksto{
\noindent The National Science Foundation (NSF) partially supported the research under
grant \#2112533: ``NSF Artificial Intelligence Research Institute for Advances in
Optimization (AI4OPT)” and grant \#2346058: ``NRT-AI: Integrating Artificial
Intelligence and Operations Research Technologies".}

\section{Introduction}

Wildfires are a growing threat under a worsening global climate. For instance, the fire intensity potential in California is expected to significantly increase by 2050~\cite{brown2025potential,chen2022landscape}, leading to more destructive and widespread wildfires. Fires ignited by power grid equipment are among the most destructive in California~\cite{CalFire2025Top}. Wildfires ignited from power grid equipment tend to burn more area than fires ignited from other sources~\cite{syphard2015location}, likely because conditions that lead to ignitions from power equipment (dry vegetation, high temperatures, and strong winds) also contribute to faster spreading fires.

To mitigate the acute risk of wildfire ignition, utilities proactively de-energize lines in high wildfire ignition risk areas through Public Safety Power Shutoff (PSPS) events~\cite{wong2022support,vazquez2022}. While de-energizing lines eliminates their ignition risk, this can lead to load shedding when the partially de-energized system cannot supply all loads. To optimally manage wildfire ignition risk and load shedding, researchers study Optimal Power Shutoff (OPS) problems to determine line de-energization decisions~\cite{Rhodes2021Balancing,piansky2025quantifying}. Typically formulated as Mixed-Integer Linear Programs (MILP), OPS problems pose significant computational challenges that we address in this paper via machine learning (ML) guided MILP solving techniques.

Utilities consider a number of factors, such as vegetation, equipment age, and Fire Potential Indices (FPI), to assign an ignition risk to each line~\cite{PGE2026Base}.  In common risk mitigation practices, lines that surpass a risk threshold are de-energized. Relative to risk thresholds applied to each line individually, OPS formulations typically permit flexibility in managing system-wide risk simultaneously across all lines, thus resulting in de-energization decisions that yield lower load shedding compared to less flexible threshold-based decisions~\cite{piansky2025quantifying}. 

FPIs and other conditions change daily, necessitating quick responses from utilities. However, solving OPS problems with realistic large-scale power system models can require hours to days of computation time~\cite{piansky2025quantifying}, rendering the resulting solutions operationally irrelevant. Thus, new computational methods are needed to operationalize OPS formulations for practically relevant system models. 

Recent advances in ML-guided MILP solving~\cite{huang2024distributional,scavuzzo2024machine} hold significant promise for addressing the computational challenges of OPS problems. Recognizing that MILPs with similar structures (i.e., MILPs of the same problem formulation constructed from data parameters sampled from a given distribution) are solved repeatedly in many applications, this body of research focuses on using ML to guide existing algorithmic policies or build new ones customized for specific instance distributions.

Two key features of OPS problems make the application of these recent ML advances particularly attractive. First, practical implementations of OPS problems would require repeated solution of very similar problems to determine daily line de-energization decisions. With limited changes to the network and conventional generators' characteristics, differences between OPS problems on a daily basis are primarily driven by varying wildfire ignition risks, load demands, and renewable generation, all of which have temporal and spatial patterns. Second, OPS formulations inherently involve significant modeling assumptions (e.g., the use of power flow approximations~\cite{haag2024long}) and parameter uncertainties associated with forecasts for wildfire ignition risk, load demands, and renewable generator outputs~\cite{eidenshink2012united,hong2016,xu2023}. By permitting the use of shorter forecast horizons, nearly optimal decisions obtained more quickly can potentially achieve better outcomes in practice than high-precision optima that require long computing times. Current OPS solutions on large realistic test networks can take upwards of 24 hours to find an optimal solution. This is too slow  for daily operations informed by the latest risk forecasts, thus motivating the development of faster solution methods. To summarize, the limited variation across OPS problems and modest value of high-precision global optima make it worthwhile to spend computational time in offline training of ML-guided MILP solvers to speed up day-of solution times when uncertainties are realized close to real-time operations. Thus, based on these two features, recent advances in ML-guided MILP solving such as those in~\cite{nair2020solving,han2023a,huang2024contrastive,huang2023searching} have the potential to determine OPS de-energization decisions much more quickly than traditional MILP solvers while significantly outperforming traditional threshold-based de-energization methods.


Extensions to the OPS problem consider long-term infrastructure investments to mitigate wildfire ignition risks~\cite{kody2022optimizing,piansky2024long,piansky2025optimizing,Taylor2022Framework}, microgrids for maintaining service during PSPS events~\cite{Taylor2023Managing}, and methods for improving the equity of wildfire mitigation outcomes~\cite{Taylor2023Managing,pollack2025equitably}, each of which impose further computational challenges. While not all of these extensions require rapid solutions for near-real-time operations, they could still benefit from faster computation times achieved using ML-guided solvers. For instance, many methods for long-term infrastructure hardening problems use decomposition techniques that repeatedly solve similar subproblems such that time spent training ML guided solvers may be worthwhile~\cite{piansky2024long,piansky2025optimizing}.



For large-scale MILPs, a large body of ML-guided MILP solving research focuses on ML-guided primal heuristics, where the goal is to obtain high-quality solutions within a short amount of time, instead of proving optimality. These approaches typically learn to predict the near-optimal assignment for variables in the problem \cite{nair2020solving,han2023a,huang2024contrastive}. Since an ML model's direct prediction may be infeasible, these methods refine the prediction by solving another MILP at inference time. Two major frameworks for this refinement are Neural Diving (\textsc{ND}) \cite{nair2020solving} and Predict-and-Search (PaS) \cite{han2023a, huang2024contrastive}. \textsc{ND} fixes a subset of variables to their predicted values, creating a smaller, easier-to-solve sub-MILP. While computationally efficient, this aggressive reduction of the search space may yield highly suboptimal solutions. In contrast, PaS searches for near-optimal solutions within a neighborhood based on the prediction. PaS operates on the original MILP and has a neighborhood parameter that specifies the number of variables that are allowed to change from their prescribed assignment suggested by the ML model \cite{han2023a, huang2024contrastive}. This approach offers more flexibility to correct prediction errors but is more computationally demanding, as the problem size remains unchanged.

While ML has previously been used for other wildfire applications (see, e.g.,~\cite{Yao2022Predicting} for ML predicted ignitions and~\cite{Tung2022Data} for data-driven operation predictions under PSPS events), this paper is, to the best of our knowledge, the first to propose using ML-guided MILP methods to quickly identify high-quality line de-energization decisions. As most lines are energized in the context of OPS, many binary variables take 1 as the solution value (as opposed to 0). This creates a class imbalance in the ML-guided framework, which is uncommon in other domains where ML-guided primal heuristics such as PaS and ND have been applied. Thus, in this paper, we extend the PaS framework used in \cite{han2023a,huang2024contrastive} by introducing two separate neighborhood parameters for the number of variables that are allowed to change from their ML-prescribed assignment, one for the variables predicted to be 0 and another one for the variables predicted to be 1. This allows us to integrate domain knowledge on the number of binary variables that take 0 and 1, respectively, as their solution value in the OPS problem. Finally, we propose a variant that combines PaS and ND, where the domain-informed PaS constraint acts as a safety net to restore feasibility. Results on a realistic synthetic test system of the California transmission grid~\cite{taylor2023california} show that our proposed ML-guided method significantly outperforms Gurobi. 

The remainder of this paper is organized as follows. Section~\ref{sec:methodology} details the OPS model used to generate the training set used for the ML model in Section~\ref{sec:ml}. Section~\ref{sec:case_study} details the specific power system studied in this paper to inform the training data set. Section~\ref{sec:results} shows results from the ML model. Finally, Section~\ref{sec:conclusion} concludes the paper.

\section{Problem Formulation}\label{sec:methodology}

This paper focuses on computing transmission line de-energization decisions to mitigate wildfire ignition risk. Traditional optimization approaches to this problem scale very poorly with both network size and the number of considered time periods. Machine learning methods provide a faster way to arrive at switching decisions. To establish notation and define the power shutoff formulation, we next describe the optimization model for the optimal power shutoff problem. As a benchmark, we then provide a methodology that uses a simple threshold-based rule to determine de-energization~decisions. 

\subsection{Optimal Power Shutoff Model}\label{sec:ops}

The Optimal Power Shutoff (OPS) problem minimizes the total amount of load shed across the system as transmission lines are de-energized to mitigate the risk of wildfire ignition. In Model~\ref{model:ops}, we present a single time period version of the OPS problem derived from~\cite{piansky2025quantifying}. 

\begin{model}[h]
\caption{Optimal Power Shutoff (OPS)}
\label{model:ops}
\begin{subequations}
\vspace{-0.2cm}
\begin{align}
& \mbox{\textbf{min}} \  \sum_{n \in \mathcal{N}} p_{ls}^n + \epsilon\!\!\! \sum_{\ell \in \mathcal{L}^{\text{switch}}}z^\ell\label{eq:obj} 
\\
& \mbox{\textbf{s.t.}}  \quad \nonumber 
\\
& 0 \leqslant p_{g}^i \leqslant \overline{p}_{g}^i 
&& \hspace{-0.5em} \forall i \in \mathcal{G} \label{subeq: gen limits} 
\\
& 0 \leqslant p_{ls}^n \leqslant p_{l}^n 
&& \hspace{-0.5em} \forall n \in \mathcal{N} \label{subeq: loadshed limits} 
\\ 
& -\overline{f}^\ell z^\ell \leqslant f^\ell \leqslant \overline{f}^\ell z^\ell 
&& \hspace{-0.5em} \forall \ell \in \mathcal{L}^{\text{switch}} \label{subeq: power flow limits switching}
\\
& -\overline{f}^\ell \leqslant f^\ell \leqslant \overline{f}^\ell 
&& \hspace{-2.1em} \forall \ell \in \mathcal{L} \setminus \mathcal{L}^{\text{switch}} \label{subeq: power flow limits} 
\\
& f^\ell \! \geqslant \!\! -b^\ell(\theta^{n^{\ell, \text{fr}}} \!\!\!\! - \! \theta^{n^{\ell, \text{to}}}) \! + \! |b^\ell|\underline{\theta}(1 \! - \! z^\ell) 
&& \hspace{-0.5em} 
\forall \ell \in \mathcal{L}^{\text{switch}} \label{subeq: power flow switching 1} 
\\
& f^\ell \! \leqslant \!\! - b^\ell(\theta^{n^{\ell, \text{fr}}} \!\!\!\! - \! \theta^{n^{\ell, \text{to}}}) \! + \! |b^\ell|\overline{\theta}(1 \!- \!z^\ell) 
&& \hspace{-0.5em} 
\forall \ell \in \mathcal{L}^{\text{switch}} \label{subeq: power flow switching 2} 
\\
& f^\ell = -b^\ell(\theta^{n^{\ell, \text{fr}}} \!\!\!\! - \! \theta^{n^{\ell, \text{to}}}) 
&& \hspace{-2.1em}  \forall \ell \in \mathcal{L}\setminus\mathcal{L}^{\text{switch}} \label{subeq: power flow}
\\
& \! \sum_{\ell \in \mathcal{L}^{n, \text{fr}}} \!\!\!\! f^\ell \! - \!\!\!\! \sum_{\ell \in \mathcal{L}^{n, \text{to}}} \!\!\!\! f^\ell \! = \!\!\!  \sum_{i \in \mathcal{G}^n} \!\! p_{g}^i \!- p_{l}^n \! + \! p_{ls}^n
&& \hspace{-0.5em} 
\forall n \in \mathcal{N} \label{subeq: power balance}
\\
& \! \sum_{\ell \in \mathcal{L}^{\text{switch}}} z^{\ell} R_{\ell} \leq R^{\text{PSPS}} && \hspace{-0.5em} \forall \ell \in \mathcal{L}^{\text{switch}}. \label{subeq: risk threshold}
\end{align}
\end{subequations}
\end{model}

The objective~\eqref{eq:obj} minimizes the load shed ($p^n_{ls}$) over all buses $n \in \mathcal{N}$ as well as a penalty term with parameter $\epsilon$ (selected as 0.01 per unit) to minimize the number of lines de-energized. For notational brevity, the objective function in the remainder of the paper is represented by $h(\boldsymbol{z}\mid M)$ to indicate the objective optimized for a given model $M$ dependent on the vector of de-energization decisions $\boldsymbol{z}$. Constraints~\eqref{subeq: gen limits} and~\eqref{subeq: loadshed limits} bound the generation ($p^i_g$) at each generator $i \in \mathcal{G}$ and load shed at each bus, respectively. Constraints~\eqref{subeq: power flow limits switching} and~\eqref{subeq: power flow limits} constrain the flow ($f^\ell$) on each line $\ell \in \mathcal{L}$ to be within the rated limits. We group lines at high risk of igniting a fire into the set $\mathcal{L}^\text{switch}$. Lines in this set have their rated limits modified by the binary de-energization decision variable, $z^\ell$ in~\eqref{subeq: power flow limits switching}. A value of $z^\ell = 0$ indicates that line $\ell$ is de-energized while $z^\ell = 1$ indicates a line is energized. Constraints~\eqref{subeq: power flow switching 1} and~\eqref{subeq: power flow switching 2} implement the B$\theta$ DC power flow approximation with a big-M linearization. Constraint~\eqref{subeq: power flow} models the DC power flow approximation for non-switchable lines. Constraint~\eqref{subeq: power balance} models power balance to ensure the flow of power into and out of each bus matches the power demanded ($p^n_l$) and produced at that bus. Constraint~\eqref{subeq: risk threshold} ensures that the sum of risk per line ($R_\ell$) on energized lines is below an acceptable system-wide threshold ($R^{\text{PSPS}}$). 

Note that we can infer two informative bounds on the number of de-energized lines from Model~\ref{model:ops}. First, we can find the minimum number of lines that must be de-energized to be feasible, $N^\text{min}_0$. By de-energizing the $k$ riskiest lines until we meet the constraint in~\eqref{subeq: risk threshold}, we can determine the minimum number of lines that must be de-energized to have a feasible solution. Likewise, we can determine the maximum number of lines that must be de-energized to guarantee a feasible solution, $N^\text{max}_0$. Here, the $m$ least risky lines can be de-energized until the constraint is met in~\eqref{subeq: risk threshold}. The number of de-energized lines in the optimal solution must be between $N^\text{min}_0$ and $N^\text{max}_0$. This domain-specific knowledge informs the machine learning methodology discussed in Section~\ref{sec:ml}.

\subsection{Thresholded De-energization Decisions}\label{sec:threshold}

As an alternative to an MILP formulation that optimizes de-energization decisions, one could de-energize lines based solely on a pre-specified threshold on their associated wildfire ignition risk, without considering the power flows in the network. The variable $z^\ell$ is then treated as a \textit{parameter} in Model~\ref{model:ops} to mimic current PSPS approaches which primarily consider wildfire ignition risk on a line-by-line basis. Similarly, the objective \eqref{eq:obj} is reduced to only consider load shedding. This reduces the problem to a linear program (LP) which is far less computationally challenging. For a fair comparison between OPS and threshold-based de-energization decisions, we set the value of $R^{\text{PSPS}}$ to match the remaining risk in the network after de-energizing lines based on the threshold method. 


\section{Machine Learning Enhanced Optimization}\label{sec:ml}

Each instance of the OPS problem is parameterized by the wildfire risk per line, $R_\ell$, renewable generation availability impacting the upper limit of some generation, $p^i_g$, and load demand at each bus, $p^n_l$. Parameters of Model~\ref{model:ops} drawn from the same distribution result in a set of MILP instances with similar structures, which motivates the use of ML to accelerate MILP solvers. Building on existing techniques~\cite{huang2024contrastive,han2023a,nair2020solving}, this section introduces our ML-guided framework for faster solution of OPS problems. 
\subsection{Solution Prediction}\label{sec:prediction}
\subsubsection{Learning task}
We use ML to predict the solution for the set of de-energization variables, $\boldsymbol{z}=[z^\ell]_{\forall \ell \in \mathcal{L}^{\text{switch}}}$ in~\eqref{subeq: power flow limits switching}, as the complexity of MILPs significantly depends on these binary decisions. Following \cite{nair2020solving} and \cite{han2023a}, we learn the probability distribution of the solution space given a problem instance. This approach learns from a set of multiple solutions given a problem instance and considers the solution quality of each solution in the training data. Specifically, given an MILP instance $M$ of the OPS problem, let $h(\boldsymbol{z}\mid M)$ denote the objective value obtained by a solution $\boldsymbol{z}$ (computed according to \eqref{eq:obj}). A solution quality function $E(\boldsymbol{z}\mid M)$ is defined as $h(\boldsymbol{z}\mid M)$ if $\boldsymbol{z}$ is feasible or $\infty$ otherwise.
Given $M$, the conditional distribution of a solution $\boldsymbol{z}$ is modeled as 
\begin{equation}
		\begin{aligned}
			\mathbb{P}(\boldsymbol{z}\mid M) \equiv \frac{\text{exp}(-E(\boldsymbol{z}\mid M))}{\sum_{\boldsymbol{z}^\prime \in Z}\text{exp}(-E(\boldsymbol{z}^\prime\mid M))},\\
		\end{aligned}\label{eq:predict:energy}
\end{equation}
where $Z$ is the set of solutions in the solution space of $M$, so that solutions with better objective values are assigned higher probabilities. The learning task is to train an ML model  parameterized by $\omega$ that approximates $\mathbb{P}(\boldsymbol{z}\mid M)$. Specifically, we learn a policy $\pi(\boldsymbol{z}\mid \omega,M)$ that outputs a prediction for the solution value for each variable in $\boldsymbol{z}$, given $M$ as the input. The policy $\pi(\boldsymbol{z}\mid \omega,M)$ produces the probabilities for which the binary variables $\boldsymbol{z}$ in Model~\ref{model:ops} take values of 1 in the solution. The output for a variable $z^\ell$ should be close to 1 if its value in the optimal solution is 1, and close to 0 otherwise.

\subsubsection{Loss function}

Following \cite{nair2020solving} and \cite{han2023a}, we use Weighted Cross-Entropy loss \cite{han2023a} to learn the policy $\pi(\boldsymbol{z}\mid\omega,M)$. We collect a training dataset that contains $N$ MILPs instances $\left\{\left(M_i,\;S_{M_i}\right)\right\}^N_{i=1}$, where $S_{M_i}$ is a set of high-quality solutions for the instance $M_i$. 
Let $\pi\left(\boldsymbol{z}\mid \omega, M_i\right)$ denote the probability of solution $\boldsymbol{z}$ given instance $M_i$ as the input. Based on the Kullback-Leibler divergence which measures the distance between the conditional distribution in (\ref{eq:predict:energy}) and the learned policy, the loss function to be minimized is:
\begin{equation}
		\begin{aligned}
			\mathcal{L}\left(\omega\right) \equiv - \sum^N_{i=1} \sum_{\boldsymbol{z} \in \mathcal{S}_{M_i}} w(\boldsymbol{z}\mid M_i)\,\text{log}( \pi\left(\boldsymbol{z}\mid  \omega;M_i\right))\\
		\end{aligned}\label{eq:kl},
\end{equation}
where 
$w(\boldsymbol{z}\mid M_i) \equiv 
\frac{
    \exp\left(-h_i(\boldsymbol{z}\mid M_i)\right)
}{
    \sum_{\boldsymbol{z}^\prime \in \mathcal{S}_{M_i}}
    \exp\left(-h_i(\boldsymbol{z}^\prime \mid M_i)\right)
}$
is the weight applied to the solution $\boldsymbol{z}$ for instance $M_i$.
Following prior work, we assume conditional independence between variables to make the training task tractable.
Given an instance $M_i$, let $\boldsymbol{z}_{l}$ denote the $l^{\text{th}}$ element 
of a solution vector $\boldsymbol{z}$.
Conditional independence leads to 
$\pi(\boldsymbol{z}\mid \omega,M)
= \prod_{l=1}^{|\boldsymbol{z}|}
\pi\left(\boldsymbol{z}_{l}\mid \omega,M\right).$

\subsection{Neural Architecture}
To learn the policy $\pi(\boldsymbol{z}\mid \omega,M)$, we represent an input MILP instance as a graph and then pass the graph into a Graph Attention Network (GAT)~\cite{brody2021attentive}, following existing literature on ML-guided primal heuristics \cite{nair2020solving,han2023a,huang2024contrastive}. We convert an MILP instance $M$ into a featured graph $\Gamma = ( \mathcal G, V, C, E)$ following~\cite{gasse2019exact}, where \(\mathcal G = (\mathcal V, \mathcal C, \mathcal E)\) is a bipartite graph that contains two sets of nodes: the variable nodes \(\mathcal V = \{1,\dots,|\mathcal{L}^{\text{switch}}|\}\) representing the decision variables \(\boldsymbol{z}\), and the constraint nodes \(\mathcal C\) representing the set of all constraints in Model~\ref{model:ops}. The edges \(\mathcal E\) represent the non-zero entries in the constraint-coefficient matrix associated with Model~\ref{model:ops}, connecting the nodes in $\mathcal V$ and $\mathcal C$. The matrices \(V\in\mathbb{R}^{|\mathcal V|\times v'}\), \(C\in\mathbb{R}^{|\mathcal C|\times c'}\), and \(E\in\mathbb{R}^{|\mathcal E|\times e'}\) are the feature vectors for $\mathcal V$, $\mathcal{C}$, and $\mathcal E$, respectively. We adopt the feature set from~\cite{gasse2019exact}, resulting in \(v'=15\) variable features (e.g., variable type, coefficients, bounds, root‐LP statistics), \(c'=4\) constraint features (e.g., constant term, sense), and \(e'=1\) edge feature (coefficient value). 

The GAT module processes the graph $\Gamma$ before outputting the predictions $\pi(\boldsymbol{z}\mid \omega,M)$. It first uses learned linear projections to embed the original features ($V,C,E$) into a common \(K\)-dimensional space, yielding 
\[
V^1 \in \mathbb{R}^{n\times K},\quad
C^1 \in \mathbb{R}^{m\times K},\quad
E^1 \in \mathbb{R}^{|\mathcal{E}|\times K}.
\]
The GAT module then performs two rounds of message passing: in round one, each constraint node in \(C^1\) attends over its incident edges in $\Gamma$ to produce updated constraint embeddings \(C^2\); in round two, each variable node in \(V^1\) attends over its incident edges to produce updated variable embeddings \(V^2\). The message passing module is followed by a multi‑layer perceptron with a sigmoid activation function, which outputs a single value as the solution prediction for each variable $z_l$.

\subsection{Inference Time Refinement}\label{inf}
As the predictions $\pi(\boldsymbol{z}\mid \omega,M)$ can be infeasible after rounding, the predictions need to be refined at inference time. Neural Diving (ND) \cite{nair2020solving} creates the final solution by fixing a subset of variables to their predicted values and creating a smaller sub-MILP. While computationally efficient, this aggressive search space reduction may yield suboptimal solutions. Predict-and-Search (PaS) \cite{han2023a, huang2024contrastive} is another framework that searches for near-optimal solutions within a neighborhood based on the prediction. The PaS \cite{han2023a, huang2024contrastive} framework includes three parameters: $k_0$, $k_1$, and $\Delta$. Denote $\mathcal{I}_0$ as a set with $k_0$ variables that contains the indices for the binary variables with the lowest $\pi(\boldsymbol{z}_l\mid \omega,M)$ values. Similarly, $\mathcal{I}_1$ is a set with $k_1$ variables that contains the indices for the binary variables with the highest $\pi(\boldsymbol{z}_l\mid \omega,M)$ values. The scalar $\Delta$ specifies the number of variables for which the solution value of the variable is allowed to change in $\mathcal{I}_0 \cup \mathcal{I}_1$. PaS assigns all variables in $\mathcal{I}_0$ to 0 and all variables in $\mathcal{I}_1$ to 1 in the MILP but also allows $\Delta$ of the assigned variables to be flipped by the solver. Accordingly, PaS adds the following constraint to the original problem: 
\begin{equation}
B(\mathcal{I}_0,\mathcal{I}_1,\Delta) = \{\boldsymbol{z} : \sum_{z_l\in \mathcal{I}_0}z_l + \sum_{z_l \in \mathcal{I}_1}1-z_l \leq \Delta\}.
\end{equation}

\subsubsection{Domain-informed PaS (\textsc{PaS}$^{\textsc{OPS}}$)}\label{ml1}
As discussed in Section~\ref{sec:ops}, the OPS problem is infeasible when the number of zero entries in $\boldsymbol{z}$ is less than $N_0^{\text{min}}$ and the optimal solution for $\boldsymbol{z}$ has a number of zero entries between $N_0^{\text{min}}$ and $N_0^{\text{max}}$. This domain knowledge informs our choice of $k_0$ and $k_1$, as well as motivates us to break the parameter $\Delta$ into $\Delta_0$ and $\Delta_1$ as separate parameters for the number of binary variables allowed to change from their prescribed assignment from those assigned to $0$ ($\mathcal{I}_0$) and those assigned to $1$ ($\mathcal{I}_1$). We set $k_0=N_0^{\text{max}}$ (in order not to exclude the optimal solution) and $\Delta_0=N_0^{\text{max}}-N_0^{\text{min}}$ (in order not to cause infeasibility due to too many variables assigned to 0). 

Additionally, based on typical wildfire risk profiles, a large percentage of variables take values of 1  in the solution (i.e., most lines are energized). Based on this observation, we set $k_1$ as a fraction of the total number of binary variables ($k_1=\phi|\mathcal{L}^{\text{switch}}|$). We also identify a lower bound for $\Delta_1$. We know that the problem is infeasible when the number of one entries is greater than $|\mathcal{L}^{\text{switch}}|-N_0^{\text{min}}$. In this case, at least $k_1-(|\mathcal{L}^{\text{switch}}|-N_0^{\text{min}})$ variables assigned to 1 need to flip in order to maintain feasibility. Therefore, the lower bound for $\Delta_1$ is  
$\text{max}(k_1-(|\mathcal{L}^{\text{switch}}|-N_0^{\text{min}}),0)$. We introduce an offset parameter $\phi'$ so that $\Delta_1 = \text{max}(k_1-(|\mathcal{L}^{\text{switch}}|-N_0^{\text{min}}),0) + \phi'k_1$. 

\textsc{PaS}$^{\textsc{OPS}}$ has the domain-informed $k_0,\Delta_0$ and  hyperparameters $\phi,\phi'$ (which determine $k_1,\Delta_1$), and adds the following two constraints to the original MILP: 
\begin{subequations}
\begin{align}
B_0(\mathcal{I}_0,\Delta_0) & = \{\boldsymbol{z} : \sum_{z_l\in \mathcal{I}_0}z_l \leq \Delta_0\},\\
B_1(\mathcal{I}_1,\Delta_1) & = \{\boldsymbol{z} :\sum_{z_l \in \mathcal{I}_1}1-z_l \leq \Delta_1\}.
\end{align}    
\end{subequations}

Note that $\phi$ and $\phi'$ are shared hyperparameters for all instances, while $k_0$, $k_1$, $\Delta_0$, and $\Delta_1$ are instance-specific. 

\subsubsection{Domain-informed PaS+ND with safety net (\textsc{PaS}$^{\textsc{OPS}}$+\textsc{ND})}\label{ml2}
We also consider a special case where we set $\Delta_1$ to its lower bound $\text{max}(k_1-(|\mathcal{L}^{\text{switch}}|-N_0^{\text{min}}),0)$ with $\phi'=0$. 
When $\phi$ is such that $k_1\leq|\mathcal{L}^{\text{switch}}|-N_0^{\text{min}}$, then $\Delta_1=0$ and no variables in $\mathcal{I}_1$ are allowed to flip. 
This is similar to ND, where a subset of variables ($\mathcal{I}_1$) is fixed, creating a smaller sub-MILP. 
In this case, this variant can be viewed as a combination of PaS and ND, where for the 0-assignments we use the $k_0$ and $\Delta_0$ settings from \textsc{PaS$^{\text{OPS}}$}, but for the 1-assignments we use the more aggressive ND with $\phi'=0$ and $\phi \leq 1 - \frac{N_0^{\text{min}}}{|\mathcal{L}^{\text{switch}}|}$ to encode our domain-specific beliefs that we are more likely to be correct for 1-assignments as more lines will be energized. 
Additionally, for higher values of $\phi$ that make $k_1>|\mathcal{L}^{\text{switch}}|-N_0^{\text{min}}$, the PaS constraint $B_1(\mathcal{I}_1,\Delta_1)$ acts as safety net for ND that can help restore feasibility, in case setting too many variables to 1 leads to infeasibility.



\section{Case Study and Datasets}\label{sec:case_study}

\begin{figure}[b]
    \centering
	\includegraphics[width=\linewidth]{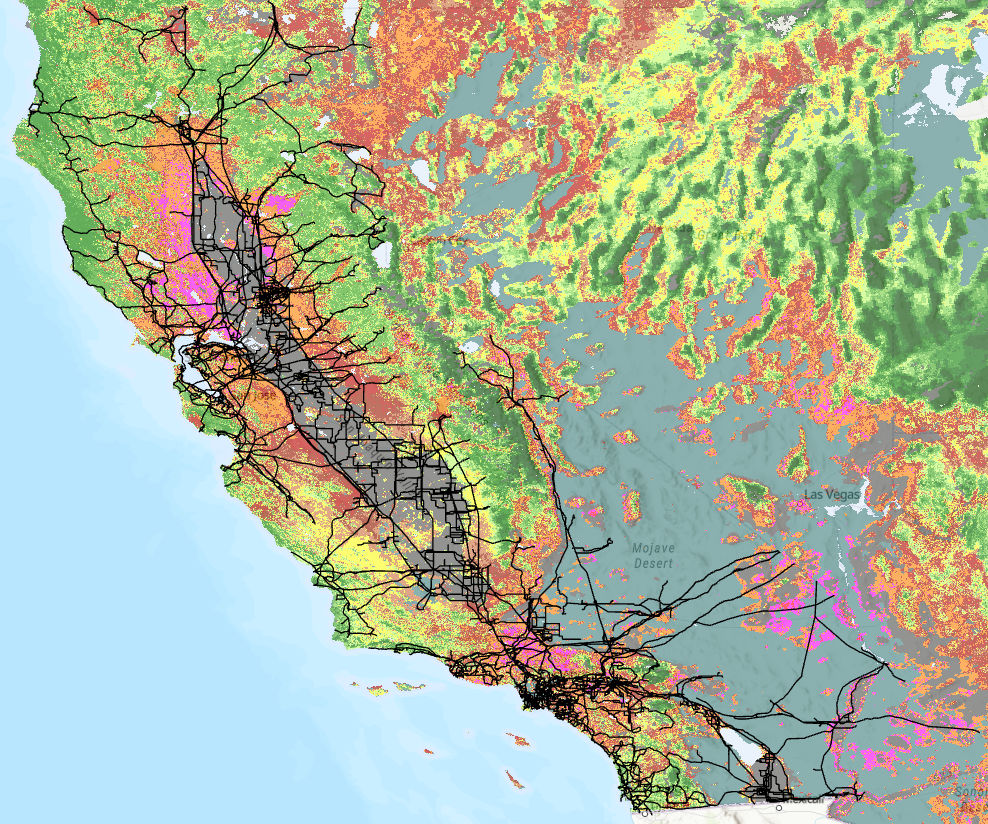}
	\caption{California's transmission line paths on a Wildland Fire Potential Index map for Oct.~26, 2020. Pixels are individually colored with lower values in green and higher values in red and pink.}
	\label{fig:cats_wfpi_network}       
\end{figure}

To demonstrate the performance of ML-guided solvers for the OPS problem in Model~\ref{model:ops}, we use the California Test System (CATS), a synthetic transmission network with line routings accurate to the actual transmission corridors in California~\cite{taylor2023california}. This network has hourly load and renewable generation data at each bus derived from real world sources providing a realistic-but-not-real synthetic transmission network. This network was augmented in~\cite{piansky2025quantifying} with daily real world wildfire ignition risk data derived from the US Geological Survey's (USGS) Wildland Fire Potential Index (WFPI)~\cite{USGS2024Wildland}. The USGS WFPI provides a 1~km by 1~km map of the US on each day with wildfire potential values ranging from 0 (low risk) to 247 (high risk). Each transmission line can intersect multiple pixels. Figure~\ref{fig:cats_wfpi_network} shows the CATS network with USGS WFPI risk information.

Reference~\cite{piansky2025quantifying} shows that the choice of wildfire risk aggregation method can impact the network topology resulting from an OPS solution. For the purpose of highlighting the benefits of our machine learning enhanced solution methodology, we focus on a single metric (maximum intersecting risk for each line) in this work. We also consider a single time period instance of the OPS problem. Specifically, we consider the hour with the most load shed from the thresholded version of the problem run on each hour of the day as defined in~\cite{piansky2025quantifying} and discussed in Section~\ref{sec:threshold}. This threshold is based on the 95\textsuperscript{th}-percentile of pixels intersected by all lines in the network over 2019 as defined in~\cite{piansky2025quantifying}. After determining the hour of the day with the most load shed, we solve the OPS problem to reach the same amount of acceptable risk while minimizing the amount of load shed. Some days have no load shed or no wildfire risk so we do not solve an OPS problem on each day. We solve these single-hour OPS problems for 2020 and 2021, resulting in 651 optimization problems for use in our ML methodology. Since the wildfire risk profile is modeled as constant throughout the day, focusing on the hour with the most load shed allows for the optimizer to prioritize mitigation of the worst-case time period.

These OPS problems are solved using Gurobi~12.0.2~\cite{gurobi}. We limit the computing time to 24-hours for each problem and save the ten incumbent solutions with the best objective values encountered through the solution process for use in the ML training as discussed in Section~\ref{sec:prediction}. 


For the ML method described in Section~\ref{sec:ml}, we classify the difficulty of each daily problem instance as either ``easy'' or ``hard''. Easy problems can be solved using conventional MILP solvers such as Gurobi in a reasonable amount of time (on the order of an hour). Hard problems under this classification tend to take hours to days of computing time to reach an optimal solution, making them a good target for ML-guided optimization techniques to provide high-quality switching decisions on a day-ahead operational time scale. 

We determine an easy/hard classification based on the amount of load shed seen in the thresholded problem since, as an LP, this problem can be solved within several minutes. If the thresholded de-energization problem yields relatively low load shed (less than 100~MWh based on our prior experience in~\cite{piansky2025quantifying}), the OPS problem is typically able to find an optimal solution quickly. Classifying days based on the initial load shed leads to 111 easy problems and 540 hard problems. 
We apply our ML methods on problems classified as hard, as the easy problems can be solved to optimality with Gurobi within a short amount of time. We use 80\%, 10\%, and 10\% of the instances for training, validation, and test. This results in 432, 54, and 54 instances for train, validation, and test, respectively. The average number of binary variables and constraints in the instances is $74,933.72$ and $73,546.17$, respectively.

\section{Results}\label{sec:results}
This section presents our evaluation metrics, implementation setup, the results for the ML-guided optimization method alongside several benchmark methods.
\subsection{Evaluation Metrics}
We use the following metrics to evaluate the effectiveness
of different methods:
(1)~The \textit{Objective Value} (OV) is the best known objective value of the MILP found by the time cutoff. (2)~The \textit{Primal Gap} (PG) \cite{berthold2013measuring} is the normalized difference between the objective value $v$ found by a method and a best known objective value $v^*$, defined as $\text{PG} = \frac{|v-v^*|}{\max(|v|,|v^*|)}$, when $vv^*>0$. When no feasible solution is found or when $vv^* <0$, PG is defined to be 1. PG is 0 when $|v|=|v^*|=0$. (3)~The \textit{Primal Integral} (PI) \cite{berthold2013measuring} is the integral of the primal gap over time, which captures the speed at which better solutions are found. (4)~The \textit{number of wins in terms of PI} (\# wins) is the number of MILP instances in the test set for which the method results in the lowest PI across all other methods. 
Note that lower values for the objective, primal gap, and primal integral indicate superior performance.

\subsection{Implementation Setup}
We use the PyTorch framework to train the ML model, and training is done on an NVIDIA A100 GPU with 128~GB of memory. We use the Adam optimizer \cite{adam2014method} with a learning rate of $0.0001$ and trained the model for 6000 epochs, which took 26.32 hours. At inference time, we compare our ML-guided approaches with Gurobi and set the time limit to 1,800 seconds for each MILP instance for all methods. We conduct testing on a cluster with epyc-7542 CPUs with 64 GB RAM. To find the hyperparameters $\phi$ and $\phi'$ for the ML guided approaches in Section~\ref{inf}, we experiment with $\phi \in \{0.5,0.6,0.7,0.8,0.9\}$ and $\phi' \in \{0.05, 0.1, 0.15,0.2,0.25,0.3,0.35\}$. We select the values that result in the lowest PI on the validation set and then apply the selected values to the test set at inference time. For \textsc{PaS}$^{\textsc{OPS}}$ the final hyperparameters used are $\phi=0.9$ and $\phi'=0.05$. For \textsc{PaS}$^{\textsc{OPS}}$+\textsc{ND}, $\phi=0.7$ is used.

\subsection{Results and Discussions}
\begin{figure}[h]
    \centering
    \includegraphics[width=0.6\columnwidth]{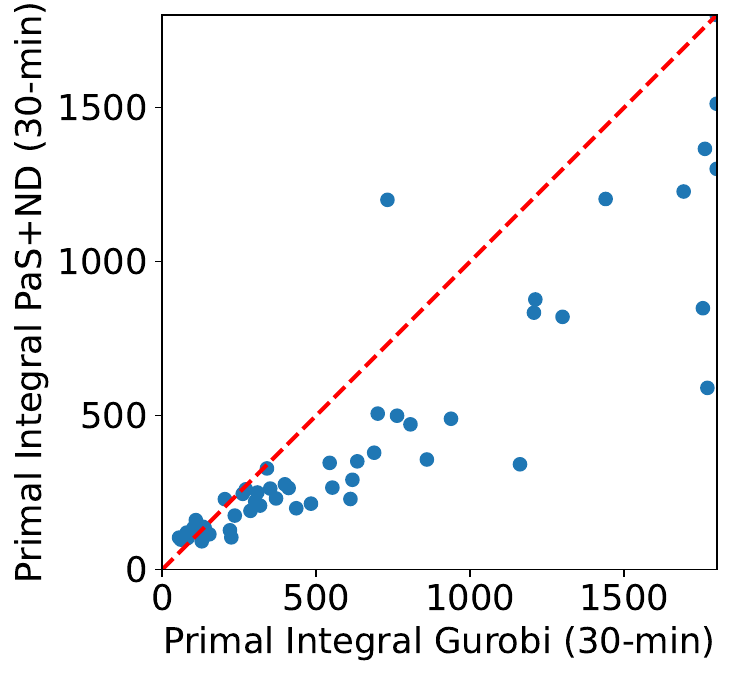}
    \caption{Scatter plot of primal integral with Gurobi vs primal integral with \textsc{PaS$^{\textsc{OPS}}$+ND}.}
    \label{fig:scatter}
\end{figure}

\begin{figure}[h]
    \centering
    \includegraphics[width=0.7\columnwidth]{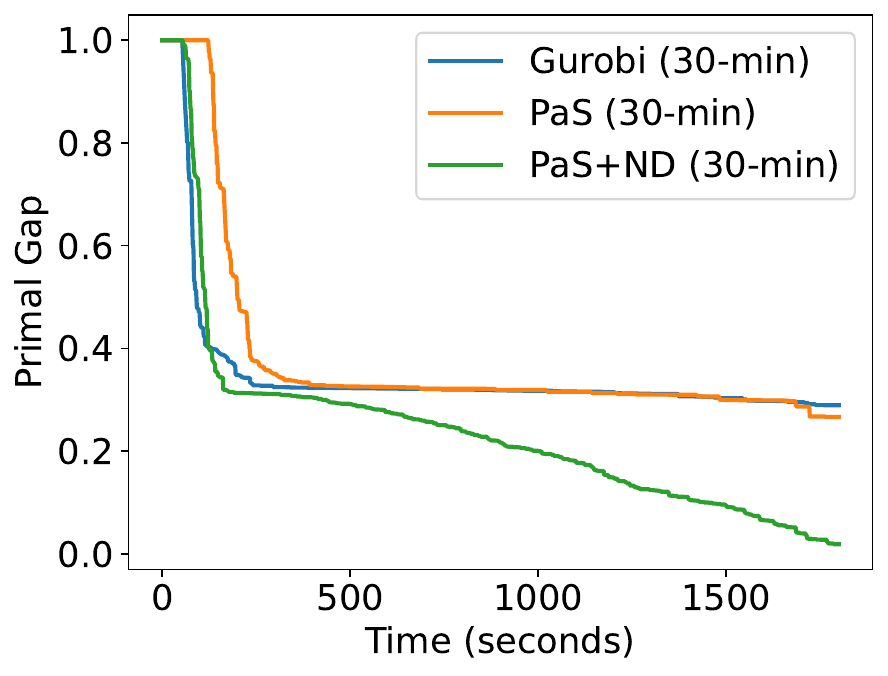}
    \caption{Average primal gap as a function of time (seconds) for Gurobi, \textsc{PaS}$^{\textsc{OPS}}$, and \textsc{PaS}$^{\textsc{OPS}}$+\textsc{ND}.}
    \label{fig:pg}
\end{figure}

We compare the proposed ML guided approaches (\textsc{PaS}$^{\textsc{OPS}}$ and \textsc{PaS}$^{\textsc{OPS}}$+\textsc{ND}) against Gurobi. All the three methods are able to find feasible solutions for all instances in the test set. As shown in Table~\ref{tab:results}, \textsc{PaS}$^{\textsc{OPS}}$+\textsc{ND} performs the best among the three methods in all the evaluation metrics considered, resulting in the lowest average objective value, primal gap, and primal integral and the highest number of wins in terms of primal integral on the test instances. \textsc{PaS}$^{\textsc{OPS}}$+\textsc{ND} significantly outperforms Gurobi, reducing the primal integral by $30.94\%$ relative to Gurobi. Figure~\ref{fig:scatter} shows a scatter plot of the primal integral with \textsc{PaS}$^{\textsc{OPS}}$+\textsc{ND} and with Gurobi, where each point represents an MILP instance in the test set. Many points in the scatter plot lie below the 45-degree line, indicating that \textsc{PaS}$^{\textsc{OPS}}$+\textsc{ND} often outperforms Gurobi. Notably, the biggest gains of \textsc{PaS}$^{\textsc{OPS}}$+\textsc{ND} are on the hardest instances. Additionally, Figure~\ref{fig:pg} shows the primal gap, averaged over the test set, as a function of time. \textsc{PaS}$^{\textsc{OPS}}$+\textsc{ND} produces high-quality solutions at a faster speed across most time steps compared to Gurobi.

\textsc{PaS}$^{\textsc{OPS}}$ performs worse than \textsc{PaS}$^{\textsc{OPS}}$+\textsc{ND} and Gurobi in most of the metrics considered. A possible explanation is that \textsc{PaS}$^{\textsc{OPS}}$ solves an MILP that contains the same number of binary variables as the original MILP at inference time, as discussed in Section~\ref{inf}. Therefore, it takes a longer time for \textsc{PaS}$^{\textsc{OPS}}$ to produce solutions with lower objective values, although it provides more flexibility to correct errors from the ML predictions by solving a larger MILP, compared to \textsc{PaS}$^{\textsc{OPS}}$+\textsc{ND}. The success of \textsc{PaS}$^{\textsc{OPS}}$+\textsc{ND} on this problem can likely be attributed to the high \textit{recall} value of the ML prediction, which is the ratio of true positives to all actual positives. The \textit{recall} of the ML model with respect to the best ground truth solution is $98.99\%$, meaning that a high number of variables that take 1 as the solution value in the ground truth data are correctly predicted to 1 by the ML model. Therefore, aggressively fixing a large percentage of variables that has high prediction scores leads to good performance.


\begin{table}[t]
\caption{
Results of ML-guided methods vs Gurobi. Average Objective Value (OV), Primal Gap (PG), Primal Integral (PI), and number of wins in terms of PI (\# wins) across 54 hard test instances.  
}
\label{tab:results}
\centering
\begin{tabular}{llllll}
\toprule
Method & OV     & PG & PI & \# wins \\
\midrule
Gurobi (30-min)            & 29.60    & 0.29	      & 627.63		   &  11 \\
\textsc{PaS}$^{\textsc{OPS}}$ (30-min)       & 	29.66         & 0.27	    & 686.91   & 1   \\
\textsc{PaS}$^{\textsc{OPS}}$+ND (30-min) & \textbf{15.14}          & \textbf{0.02}	      & \textbf{433.46} & \textbf{42}   \\
\bottomrule
\end{tabular}
\end{table}

\subsection{PSPS Outcomes}
\begin{figure}[h]
    \centering
	\includegraphics[width=\linewidth]{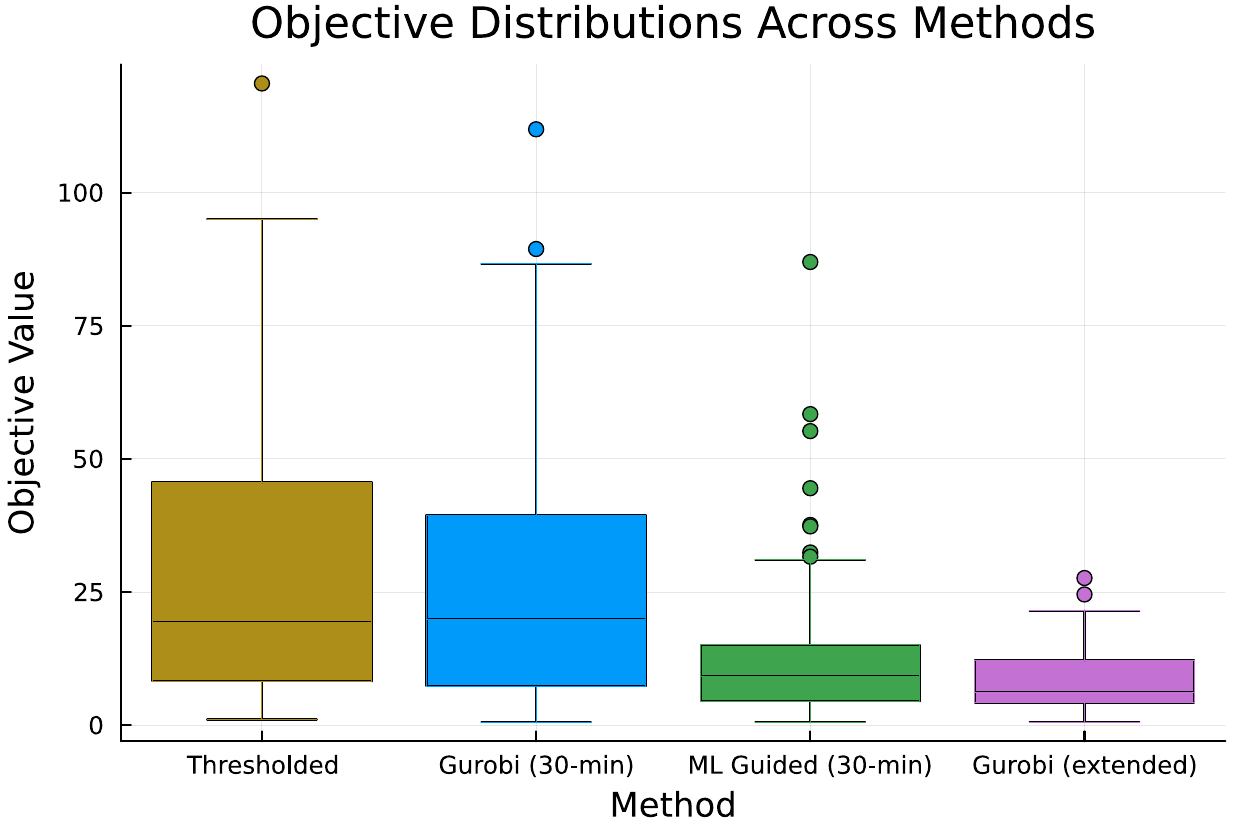}
	\caption{Box plots showing the distribution of objective outcomes across the four considered solution methodologies. Here, we only include objective values in the distribution from dates that fall into the hard test data set discussed in Section~\ref{sec:case_study}.}
	\label{fig:obj_box_plots}
\end{figure}
To better characterize the physical impacts of the solutions from each solution approach, we detail results from the thresholded version of this problem (denoted as \emph{threshold}, see Section~\ref{sec:threshold}), the ML guided outcomes at a 30-minute time limit (denoted as \emph{ML-guided 30-min}, see Section~\ref{sec:ml}), the Gurobi output at a 30-minute time limit (denoted as \emph{Gurobi 30-min}), and extended results at a 24-hour time limit or optimality at a 1\% MIP gap (denoted as \emph{Gurobi extended}, see Section~\ref{sec:ops}). For the ML guided results in this section, we consider the \textsc{PaS$^{\textsc{OPS}}$+ND} outcomes as this ML methodology achieves better performance, as shown in Section \ref{sec:results}. 

\begin{figure*}[!htbp]
    \centering
    \begin{subfigure}[t]{0.48\textwidth}
        \centering
        \includegraphics[width=\textwidth]{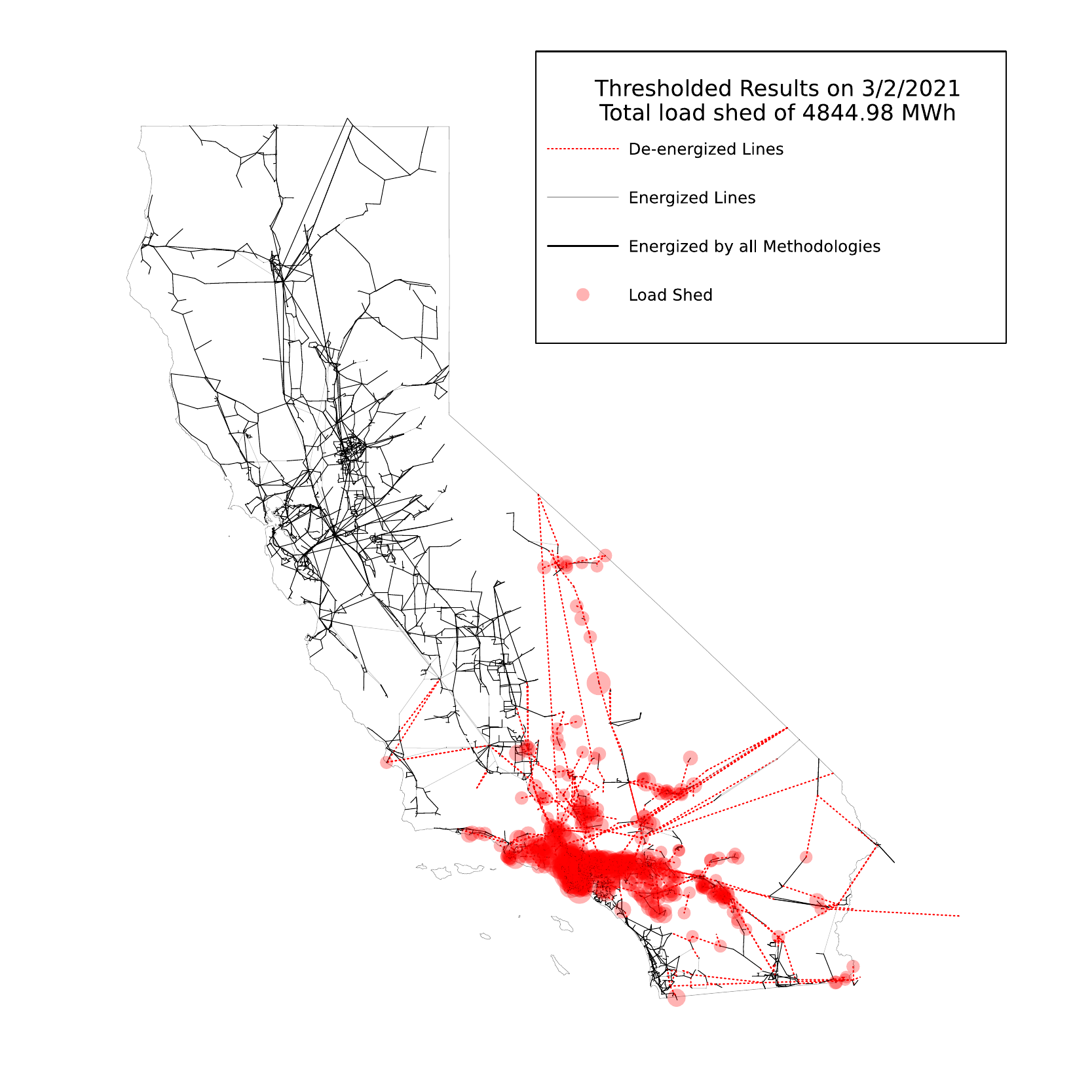}
        \caption{Thresholded line statuses.}
        \label{fig:threshold_switching}
    \end{subfigure}
    \hfill
    \begin{subfigure}[t]{0.48\textwidth}
        \centering
        \includegraphics[width=\textwidth]{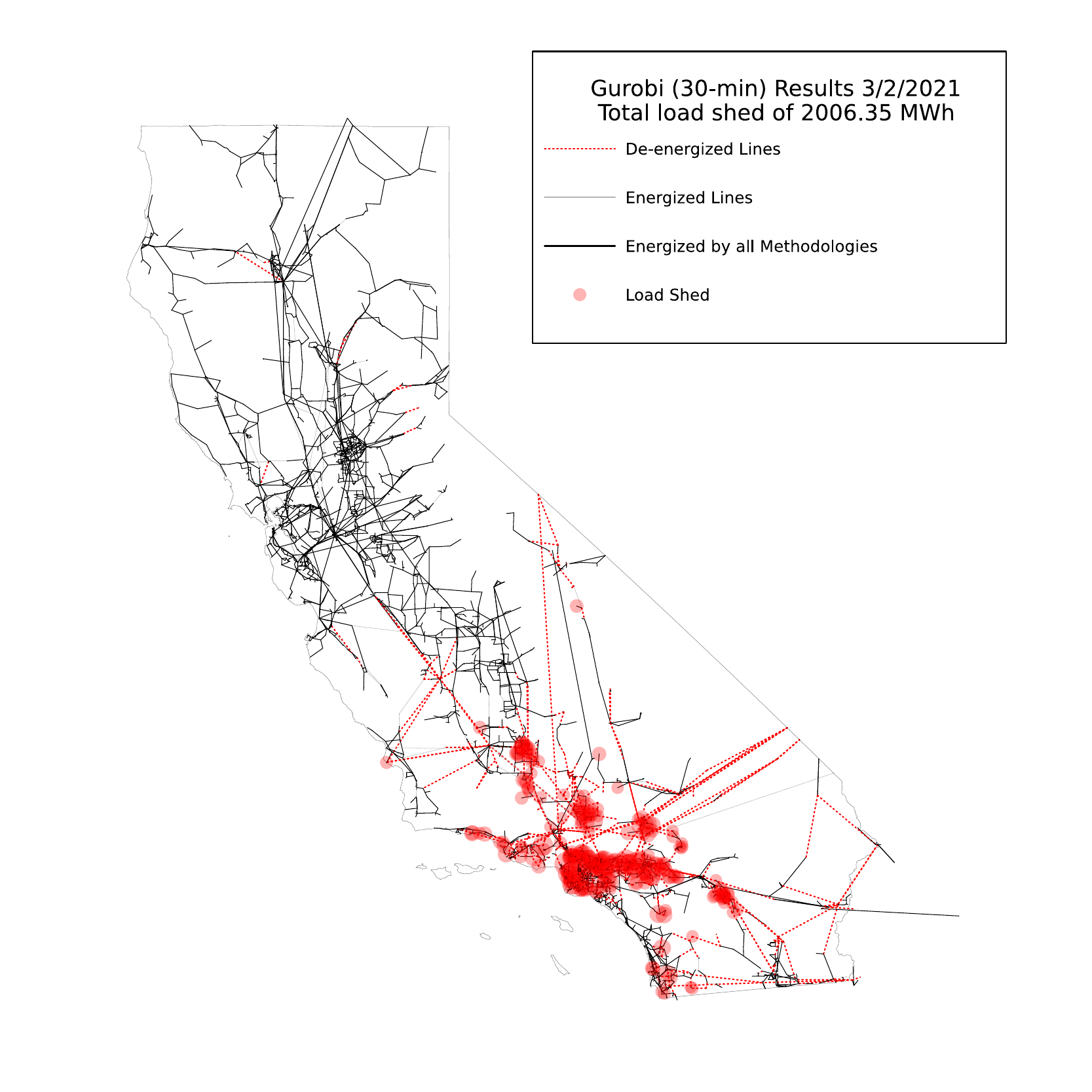}
        \caption{Line statuses from the Gurobi solution at 30 minutes.}
        \label{fig:gurobi30_switching}
    \end{subfigure}

    \vspace{0.5em}
    \begin{subfigure}[t]{0.48\textwidth}
        \centering
        \includegraphics[width=\textwidth]{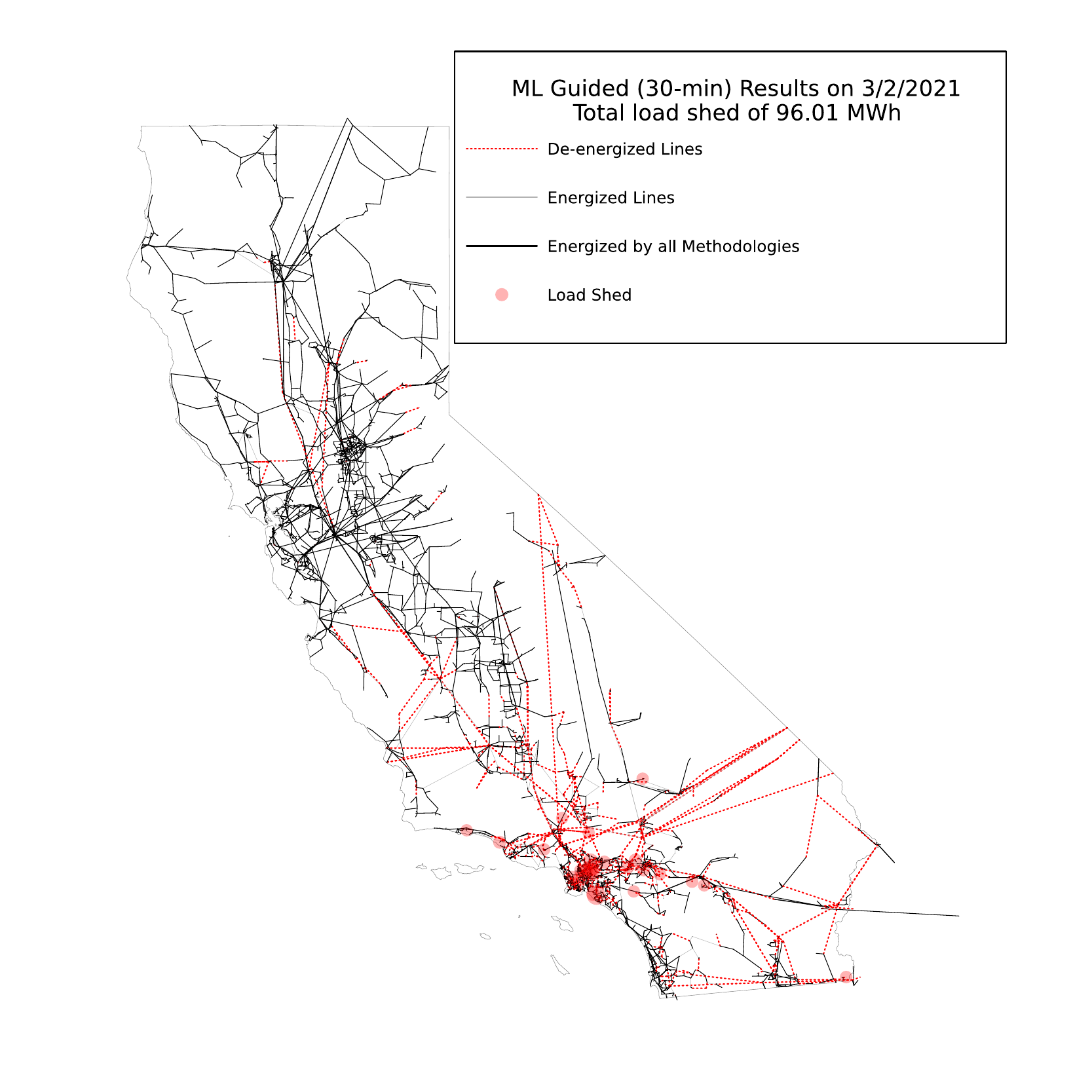}
        \caption{Line statuses from the ML-guided solution at 30 minutes.}
        \label{fig:MLguided_switching}
    \end{subfigure}
    \hfill
    \begin{subfigure}[t]{0.48\textwidth}
        \centering
        \includegraphics[width=\textwidth]{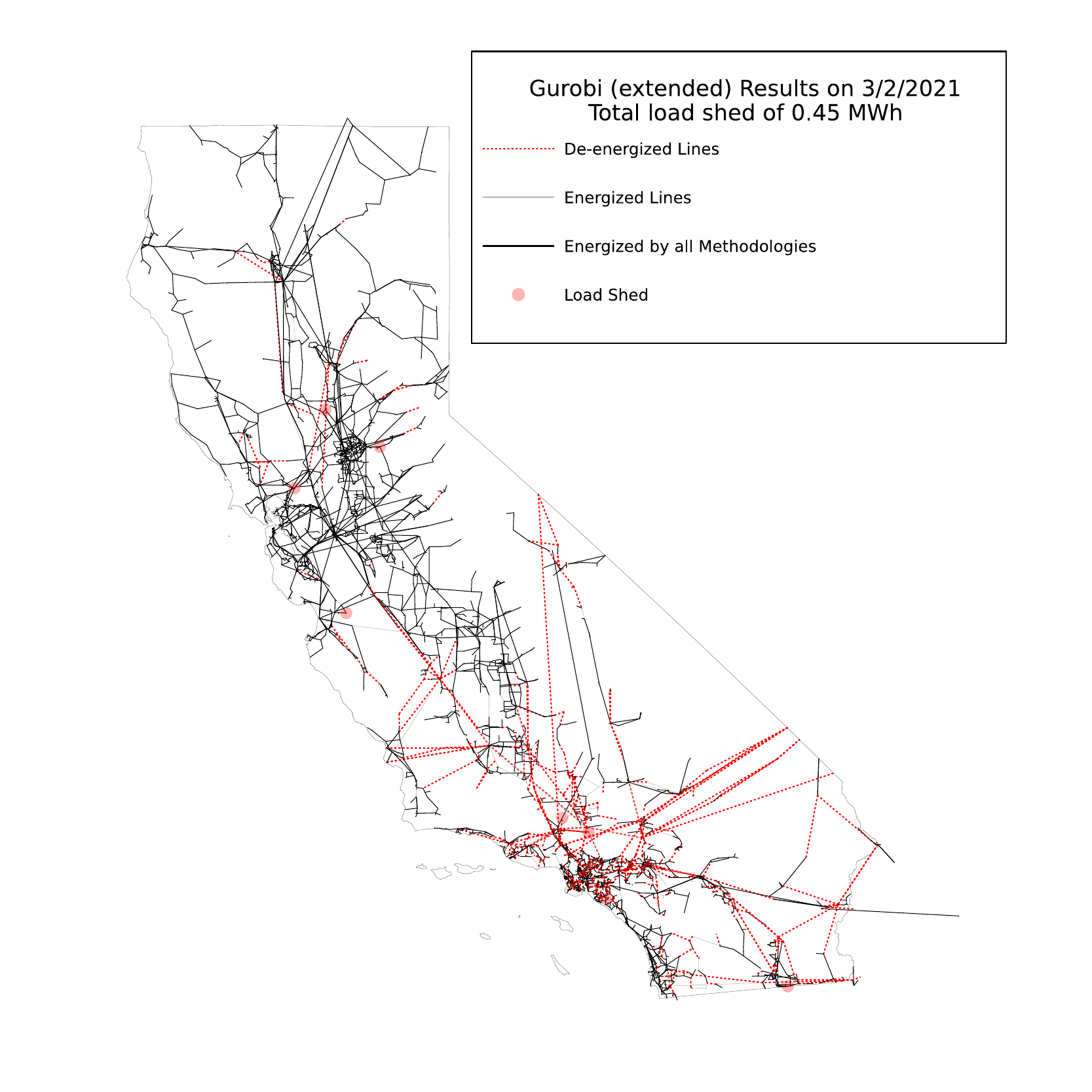}
        \caption{Line statuses from Gurobi solution at 24 hours}
        \label{fig:gurobi24_switching}
    \end{subfigure}

    \caption{Geographic plots showing the line statuses and load shed from different solution methodologies for March 2\textsuperscript{nd}, 2021, one of the hard instances. Red dashed lines show lines de-energized under that solution methodology. Black solid lines show lines that are energized across all solution methodologies for this day. Grey lines show lines that are energized in that specific methodology. Red circles indicate load shed at a bus $n$ with the size of the circle corresponding to the amount of load shed.}
    \label{fig:switching_decisions}
\end{figure*}

In Figure~\ref{fig:switching_decisions}, we show the switching outcomes from these four different solution methodologies on a representative day (March 2, 2021) with high wildfire ignition risk in southern California. We can see that the switching decisions from the Gurobi solution at a 30-minute time limit (Figure~\ref{fig:gurobi30_switching}) are similar to those from the thresholded version (Figure~\ref{fig:threshold_switching}) with large amounts of load shed. The ML-guided outcomes (Figure~\ref{fig:MLguided_switching}) have different switching decisions with much lower load shed when compared to the thresholded and 30-minute Gurobi solutions. The extended Gurobi solution, run to a 24-hour time limit (3.23\% MIP gap), achieves much lower load shed with a different network topology. Across the network, there are a large number of lines that remain energized under all solution methodologies, highlighting that the choice of line energization status can greatly impact the amount of load shed. 

The box plots in Figure~\ref{fig:obj_box_plots} show the distribution of objective values across the four different methodologies. The thresholded and 30-minute Gurobi results have similar values with the 30-minute Gurobi solutions having a slightly lower objective on average. The ML-guided and extended Gurobi solutions have similar distributions of objective values. While the extended Gurobi does reach more optimal solutions on average, this requires a 24-hour time limit as opposed to the ML guided solutions which are only run for 30 minutes.

\section{Conclusions}\label{sec:conclusion}

This paper has proposed a domain-informed ML-guided solution method for OPS problems. Compared to Gurobi, the proposed \textsc{PaS$^{\textsc{OPS}}$+ND} method finds high-quality de-energization decisions at a faster speed and reduces the primal integral by 30.94\% and primal gap by 93.10\% at the 30-minute computational time cutoff. \textsc{PaS$^{\textsc{OPS}}$+ND} also achieves large reductions in load shedding compared to traditional optimization methods with the same amount of time.  

The results in this paper demonstrate that ML-guided MILP methods have significant potential to improve solution speed for important power grid resilience problems. In the context of mitigating wildfire ignition risks, future work aims to extend this work by considering more sophisticated problem formulations that incorporate alternative power flow models~\cite{haag2024long}, multiple time periods~\cite{yang2024multi}, restoration models~\cite{rhodes2023co}, and security constraints~\cite{rhodes2023}. While no steps are taken in this work to prevent islanding or maintain critical loads, further extensions could better model isolated buses~\cite{jeanson2025scalable}. ML-guided MILP methods, especially algorithms based on large-neighborhood search~\cite{song2020general,huang2023searching}, also have potential applications in OPS-based long-term planning problems that optimize infrastructure investment decisions under wildfire ignition risk~\cite{kody2022optimizing} and other hazards~\cite{piansky2025powertech}, particularly when decomposition methods yield many closely related subproblems~\cite{piansky2024long,piansky2025optimizing}.




%
\bibliographystyle{IEEEtran}
\IEEEtriggeratref{27}
\bibliography{refs.bib}

@article{huang2024distributional,
  title={Distributional {MIPLIB}: A multi-domain library for advancing {ML}-guided {MILP} methods},
  author={Huang, Weimin and Huang, Taoan and Ferber, Aaron M and Dilkina, Bistra},
  journal={arXiv:2406.06954},
  year={2024}
}

@article{berthold2013measuring,
  title={Measuring the impact of primal heuristics},
  author={Berthold, Timo},
  journal={Operations Research Letters},
  volume={41},
  number={6},
  pages={611--614},
  year={2013},
  publisher={Elsevier}
}

@article{adam2014method,
  title={Adam: A method for stochastic optimization},
  author={Diederik P. Kingma and Jimmy Ba},
  journal={arXiv:1412.6980},
  number={},
  year={2014}
}

@article{gasse2019exact,
  title={Exact combinatorial optimization with graph convolutional neural networks},
  author={Gasse, Maxime and Ch{\'e}telat, Didier and Ferroni, Nicola and Charlin, Laurent and Lodi, Andrea},
  journal={Advances in neural information processing systems},
  volume={32},
  year={2019}
}

@article{brody2021attentive,
  title={How attentive are graph attention networks?},
  author={Brody, Shaked and Alon, Uri and Yahav, Eran},
  journal={arXiv:2105.14491},
  year={2021}
}

@article{scavuzzo2024machine,
  title={Machine learning augmented branch and bound for mixed integer linear programming},
  author={Scavuzzo, Lara and Aardal, Karen and Lodi, Andrea and Yorke-Smith, Neil},
  journal={Mathematical Programming},
  pages={1--44},
  year={2024},
  publisher={Springer}
}

@inproceedings{
han2023a,
title={A {GNN}-Guided Predict-and-Search Framework for Mixed-Integer Linear Programming},
author={Qingyu Han and Linxin Yang and Qian Chen and Xiang Zhou and Dong Zhang and Akang Wang and Ruoyu Sun and Xiaodong Luo},
booktitle={11th International Conference on Learning Representations (ICLR)},
year={2023},
}

@inproceedings{huang2024contrastive,
  title={Contrastive Predict-and-Search for Mixed Integer Linear Programs},
  author={Huang, Taoan and Ferber, Aaron M and Arman Zharmagambetov and Tian, Yuandong and Dilkina, Bistra},
  booktitle={41st International Conference on Machine Learning (ICML)},
  year={2024},
  month={July}
}

@article{nair2020solving,
  title={Solving mixed integer programs using neural networks},
  author={Nair, Vinod and Bartunov, Sergey and Gimeno, Felix and Von Glehn, Ingrid and Lichocki, Pawel and Lobov, Ivan and O'Donoghue, Brendan and Sonnerat, Nicolas and Tjandraatmadja, Christian and Wang, Pengming and others},
  journal={arXiv:2012.13349},
  year={2020}
}

@article{piansky2025quantifying,
    author={R. Piansky and S. Taylor and N. Rhodes and D. K. Molzahn and L. A. Roald and Jean-Paul Watson},
    journal={58th Hawaii International Conference on System Sciences (HICSS)}, 
    title={Quantifying Metrics for Wildfire Ignition Risk from Geographic Data in Power Shutoff Decision-Making},
    year={2025},
    volume={},
    number={},
    month={January},
    pages={}
}

@article{taylor2023california,
  title={California test system {(CATS)}: A geographically accurate test system based on the {C}alifornia grid},
  author={Taylor, Sofia and Rangarajan, Aditya and Rhodes, Noah and Snodgrass, Jonathan and Lesieutre, Bernie and Roald, Line A},
  journal={IEEE Transactions on Energy Markets, Policy and Regulation},
  year={2024},
  volume={2},
  number={1},
  pages={107-118},
}

@article{brown2025potential,
  title={The potential for fuel reduction to reduce wildfire intensity in a warming California},
  author={Brown, Patrick T and Strenfel, Scott J and Bagley, Richard B and Clements, Craig B},
  journal={Environmental Research Letters},
  volume={20},
  number={2},
  year={2025},
  publisher={IOP Publishing}
}

@misc{CalFire2025Top,
  author       = {{California Department of Forestry and Fire Protection (CalFire)}},
  title        = {Top 20 Destructive California Wildfires},
  year         = {2025},
  month        = {July},
  url          = {https://34c031f8-c9fd-4018-8c5a-4159cdff6b0d-cdn-endpoint.azureedge.net/-/media/calfire-website/our-impact/fire-statistics/top20_destruction_072525.pdf}
}

@article{syphard2015location,
  title={Location, timing and extent of wildfire vary by cause of ignition},
  author={Syphard, Alexandra D and Keeley, Jon E},
  journal={International Journal of Wildland Fire},
  volume={24},
  number={1},
  pages={37--47},
  year={2015},
  publisher={CSIRO}
}

@article{wong2022support,
  title={Support for public safety power shutoffs in {C}alifornia: Wildfire-related perceived exposure and negative outcomes, prior and current health, risk appraisal and worry},
  author={Wong-Parodi, Gabrielle},
  journal={Energy Research \& Social Science},
  volume={88},
  pages={102495},
  year={2022},
  publisher={Elsevier}
}

@inproceedings{kody2022optimizing,
	author={A. Kody and R. Piansky and D. K. Molzahn},
	booktitle={11th IREP Symposium on Bulk Power System Dynamics and Control},
	title={Optimizing Transmission Infrastructure Investments to Support Line De-energization for Mitigating Wildfire Ignition Risk},
	volume = {},
	number = {},
	pages = {},
	month = {July},
	year = {2022},
}

@ARTICLE{Rhodes2021Balancing,
      author={Rhodes, Noah and Ntaimo, Lewis and Roald, Line},
      journal={IEEE Transactions on Power Systems}, 
      title={Balancing Wildfire Risk and Power Outages Through Optimized Power Shut-Offs}, 
      year={2021},
      volume={36},
      number={4},
}

@article{piansky2024long,
	author={R. Piansky and G. Stinchfield and A. Kody and D. K. Molzahn and J. P. Watson},
	journal={Electric Power Systems Research},
	year = {2024},
	title={Long Duration Battery Sizing, Siting, and Operation Under Wildfire Risk Using Progressive Hedging},
	volume = {235},
	pages = {},
	month = {Oct},
	note ={\emph{23rd Power Systems Computation Conference (PSCC)}}
}

@article{piansky2025optimizing,
	author={R. Piansky and R. K. Gupta and D. K. Molzahn},
	title={Optimizing Battery and Line Undergrounding Investments for Transmission Systems under Wildfire Risk Scenarios: A {B}enders Decomposition Approach},
	journal={Sustainable Energy, Grids and Networks},
	year={2025},
	month={September},
	volume={43},
	number={101838},
	pages={},
	note={\textit{12th IREP Bulk Power system Dynamics and Control Symposium}}
}

@article{Taylor2022Framework, title={A framework for risk assessment and optimal line upgrade selection to mitigate wildfire risk}, volume={213},  journal={Electric Power Systems Research}, author={Taylor, Sofia and Roald, Line A.}, year={2022}, 
  note = {\emph{22nd Power Systems Computation Conference (PSCC)}},
  }

@inproceedings{Taylor2023Managing,
    title = {Managing Wildfire Risk and Promoting Equity through Optimal Configuration of Networked Microgrids},
    author = {Sofia Taylor and Gabriela Setyawan and Bai Cui and Ahmed Zamzam and Line A. Roald},
    booktitle = {ACM e-Energy},
    year = {2023},
   month={June},
    day = {},
    pages = {189-199}
}

@article{pollack2025equitably,
  title={Equitably allocating wildfire resilience investments for power grids—{T}he curse of aggregation and vulnerability indices},
  author={Pollack, Madeleine and Piansky, Ryan and Gupta, Swati and Molzahn, Daniel K.},
  journal={Applied Energy},
  volume={388},
  pages={125511},
  year={2025},
  publisher={Elsevier}
}

@article{Yao2022Predicting, 
title={Predicting electricity infrastructure induced wildfire risk in {C}alifornia}, 
volume={17},
number={9}, 
journal={Environmental Research Letters}, publisher={IOP Publishing}, 
author={Yao, Mengqi and Bharadwaj, Meghana and Zhang, Zheng and Jin, Baihong and Callaway, Duncan S.}, 
year={2022}, 
month={9}, 
pages={094035} 
}

@inproceedings{Tung2022Data,
  author={Hong, Wanshi and Wang, Bin and Yao, Mengqi and Callaway, Duncan and Dale, Larry and Huang, Can},    
title = {Data-Driven Power System Optimal Decision Making Strategy under Wildfire Events},
    booktitle = {55th Hawaii International Conference on System Sciences (HICSS)},
    year = {2022},
month={January}
}

@misc{USGS2024Wildland,
    author = {{U.S. Geological Survey}},
    title = {{Wildland Fire Potential Index}},
    year = 2024,
    url = {https://www.usgs.gov/fire-danger-forecast/wildland-fire-potential-index-wfpi}
}

@techreport{PGE2026Base,
  author       = {{Pacific Gas and Electric Company}},
  title        = {{PG\&E} Wildfire Mitigation Plan {R1} (2026–2028), Volume 1: {Base WMP}},
  institution  = {Pacific Gas and Electric Company},
  year         = {2026},
  month        = {February},
  number       = {Vol. 1},
  url          = {https://www.pge.com/assets/pge/docs/outages-and-safety/outage-preparedness-and-support/pge-base-wmp-vol-1.pdf},
}

@article{haag2024long,
  title={Long solution times or low solution quality: On trade-offs in choosing a power flow formulation for the optimal power shutoff problem},
  author={Haag, Eric and Rhodes, Noah and Roald, Line},
  journal={Electric Power Systems Research},
  volume={234},
  pages={110713},
  year={2024},
  note ={\emph{23rd Power Systems Computation Conference (PSCC)}}
}

@misc{gurobi,
  author = {{Gurobi Optimization, LLC}},
  title = {{Gurobi Optimizer Reference Manual}},
  year = 2025,
  url = "www.gurobi.com"
}

@ARTICLE{vazquez2022,
  author={Vazquez, Daniel A. Zuniga and Qiu, Feng and Fan, Neng and Sharp, Kevin},
  journal={IEEE Transactions on Power Systems}, 
  title={Wildfire Mitigation Plans in Power Systems: A Literature Review}, 
  year={2022},
  volume={37},
  number={5},
  pages={3540-3551},
}

@article{eidenshink2012united,
  title={United {S}tates {G}eological {S}urvey fire science: Fire danger monitoring and forecasting},
  author={Eidenshink, Jeff C and Howard, Stephen M},
  year={2012},
  journal={US Geological Survey},
  note={\href{https://www.usgs.gov/publications/united-states-geological-survey-fire-science-fire-danger-monitoring-and-forecasting}{https://www.usgs.gov/publications/united-states- geological-survey-fire-science-fire-danger-monitoring-and-forecasting}}
}

@article{hong2016,
  title={Probabilistic Electric Load Forecasting: A Tutorial Review},
  author={Hong, Tao and Fan, Shu},
  journal={International Journal of Forecasting},
  volume={32},
  number={3},
  pages={914-938},
  year={2016},
}

@ARTICLE{xu2023,
  author={Xu, Chen and Xie, Yao and Vazquez, Daniel A. Zuniga and Yao, Rui and Qiu, Feng},
  journal={IEEE Journal on Selected Areas in Information Theory}, 
  title={Spatio-Temporal Wildfire Prediction Using Multi-Modal Data}, 
  year={2023},
  volume={4},
  number={},
  pages={302-313},
}

@ARTICLE{rhodes2023,
  title={Security constrained optimal power shutoff for wildfire risk mitigation},
  author={Rhodes, Noah and Coffrin, Carleton and Roald, Line},
  journal={IET Generation, Transmission \& Distribution},
  volume={18},
  number={18},
  pages={2972-2986},
  year={2024},
}

@article{yang2024multi,
  title={Multi-period power system risk minimization under wildfire disruptions},
  author={Yang, Hanbin and Rhodes, Noah and Yang, Haoxiang and Roald, Line and Ntaimo, Lewis},
  journal={IEEE Transactions on Power Systems},
  volume={39},
  number={5},
  pages={6305-6318},
  year={2024},
}

@inproceedings{rhodes2023co,
  title={Co-optimization of power line shutoff and restoration under high wildfire ignition risk},
  author={Rhodes, Noah and Roald, Line A},
  booktitle={IEEE Belgrade PowerTech},
  pages={},
  year={2023},
}

@article{song2020general,
  title={A general large neighborhood search framework for solving integer linear programs},
  author={Jialin Song and Ravi Lanka and Yisong Yue and Bistra Dilkina},
  journal={33rd Advances in Neural Information Processing Systems (NeurIPS)},
  volume={},
  pages={20012--20023},
  year={2020}
}

@inproceedings{chen2022landscape,
  title={Landscape optimization for prescribed burns in wildfire mitigation planning},
  author={Chen, Weizhe and Sivaramakrishnan, Eshwar Prasad and Dilkina, Bistra},
  booktitle={Proceedings of the 5th ACM SIGCAS/SIGCHI Conference on Computing and Sustainable Societies},
  pages={429--438},
  year={2022}
}

@inproceedings{huang2023searching,
  title={Searching large neighborhoods for integer linear programs with contrastive learning},
  author={Huang, Taoan and Ferber, Aaron M and Tian, Yuandong and Dilkina, Bistra and Steiner, Benoit},
  booktitle={40th International Conference on Machine Learning (ICML)},
  pages={13869-13890},
  year={2023},
}

@article{piansky2025powertech,
	author={R. Piansky and D. K. Molzahn and N. D. Jackson and J. K. Skolfield},
	journal={IEEE Kiel PowerTech},
	title={Evaluating Undergrounding Decisions for Wildfire Ignition Risk Mitigation across Multiple Hazards},
	volume = {},
	number = {},
	pages = {},
	month = {June},
	year = {2025},
}

@article{jeanson2025scalable,
  title={Scalable Iterative Algorithm for Solving Optimal Transmission Switching with De-energization},
  author={Jeanson, Beno{\^\i}t and Tanneau, Mathieu and Tindemans, Simon},
  journal={arXiv:2511.13662},
  year={2025}
}

\end{document}